# Deep Baseline Network for Time Series Modeling and Anomaly Detection

1st Cheng Ge
*Alibaba Group*
Shanghai, China
eric.gc@alibaba-inc.com

2nd Xi Chen
*Alibaba Group*
Shanghai, China
chuyu.cx@alibaba-inc.com

3rd Ming Wang
*Alibaba Group*
Hangzhou, China
duchen.wm@taobao.com

4th Jin Wang
*Alibaba Group*
Hangzhou, China
jet.wangj@alipay.com

*Abstract*— Deep learning has seen increasing applications in time series in recent years. For time series anomaly detection scenarios, such as in finance, Internet of Things, data center operations, etc., time series usually show very flexible baselines depending on various external factors. Anomalies unveil themselves by lying far away from the baseline. However, the detection is not always easy due to some challenges including baseline shifting, lacking of labels, noise interference, real time detection in streaming data, result interpretability, etc. In this paper, we develop a novel deep architecture to properly extract the baseline from time series, namely Deep Baseline Network (DBLN). By using this deep network, we can easily locate the baseline position and then provide reliable and interpretable anomaly detection result. Empirical evaluation on both synthetic and public real-world datasets shows that our purely unsupervised algorithm achieves superior performance compared with state-of-art methods and has good practical applications.

## I. Introduction

A typical time series anomaly detection problem can be formulated as:

Problem 1: Given a sequence of length T, i.e., $y = [y_1, y_2, ..., y_T]$, to produce a label sequence of same length, $l = [l_1, l_2, ..., l_T]$, where $l \in \{0, 1\}$, $l = 1$ indicates an anomaly point while $l = 0$ not.

However, in many industry scenarios, business cannot wait for a detection until $T$ samples are ready. The data is collected in real time, per minute or per second for example, the detection needs to be completed once the data arrives. Then the problem becomes streaming time series detection:

Problem 2: Given a time series look-back window of length T (without label), i.e., $y = [y_1, y_2, ..., y_T]$, to produce a label $l_{T+1}$ of the latest point $y_{T+1}$, where $l_{T+1} \in \{0, 1\}$.

The second problem can benefit from the time series forecasting techniques which already has tons of literatures. The most classic method is the autoregressive integrated moving average(ARIMA[1]). In addition to point forecasting, ARIMA calculates the confidence interval which provides a probabilistic way of finding anomalies. However, the underlying math model of ARIMA is purely linear, which makes the predictive power limited. Besides, ARIMA needs to search the hyper-parameters of AR, MA and differencing parts during each estimation and each time series, leaving the algorithm very computation exhaustive and not effective in real time streaming detection with industrial big data. DeepAR[2], proposed by David Salinas et al., enables the recurrent LSTM network to model the likelihood of the forecasting points. It can use one model for all time series in the dataset. It utilizes the sequence-to-sequence architecture for forecasting. On top of the LSTM is the probabilistic dense layer, which estimates the density function of the target distribution. With the help of DeepAR, we can generate an interval estimation for the candidate value. However, many time series contains complex and shifting trends, leaving the data un-stationary, causing significant forecasting bias for the LSTM. The noise and anomalies in data also cause the LSTM difficult to train. N-Beats [3], proposed by Boris N. Oreshkin etc. in 2020, introduced a new deep neural architecture based on backward and forward residual links and a very deep stack of fully-connected layers. N-Beats has many functional stacks connected in series, each stack consists of several basic building blocks to form the complete network. In N-Beats, authors proposed the trend basis and seasonal basis which extract the trends and seasonal pattern respectively. However, for the trend extraction, N-beats only uses a single polynomial with small degree $p$ to mimic the trend. This is not adequate in practice. Firstly, trend is usually complex and often shifting. Second, the polynomial regression should be robust to noise and outliers. Vanilla N-Beats cannot tackle these two issues. With regard to traditional time series modeling methods such as STL[9][12], H-P trend filtering[7] and their modern derivatives including RobustTrend[5] and RobustSTL[6], they all take a statistical approach rather than neural network which utilizes the power of GPU, this limits their computation performance in real time detection of industrial big data. Spectral Residual[4], proposed by Ren etc., is another approach to anomaly detection. This algorithm borrows the idea in visual saliency detection domain and computes the time series spectral residuals. However, a significant drawback is the algorithm needs to estimate a few future points before detecting the current point. Simple linear extrapolating will cause significant bias in the future point estimation. The authors in [4] further proposed an improved algorithm called SR-CNN. However, this becomes a supervised algorithm which requires careful labeling.

In this paper, we stick to the basic idea of DeepAR to model the conditional likelihood of current data in an autoregressive way, but propose a new baseline estimation method. We name it Deep Baseline Network (DBLN for short). The method is purely unsupervised, meaning that history data labeling is not required. The model integrates locally weighted regression and recurrent network into its baseline block. The local regression layer locates the underlying baseline while reducing noise and outlier influence. The Q statistic in the loss function ensures critical information in the curve is extracted while model residual becomes white noise. The network outputs the point forecast with confidence interval which guides the anomaly detection and achieves superior results.

## II. METHODOLOGY

### A. Network design

Figure 1 shows the basic architecture of the deep baseline network. The network is based on an auto-regressive pattern. To detect point $y_{T+1}$, we provide a look-back window of size T: $y_1, y_2, \ldots, y_T$. Data in the look-back window is firstly fed into the baseline blocks. Inspired by N-Beats network, each baseline block has two outputs: backcast estimation and forecast estimation. The backcast estimation is subtracted from the input signal so that the corresponding baseline component is removed, the residual $r$ is further fed into the next block. Inside the baseline block, input firstly passes through a bi-LSTM layer. The bi-LSTM is used to predict the local regression coefficients for each point in the look-back window. And then, a local regression is run to output the backcast $b$ and forecast prediction $f$. The forecasts of all blocks are summed up to be the final forecast. To express it mathematically, let's denote the input to $l$-th baseline block as $z_l$, the backcast and forecast of $l$-th baseline block are $b_l$ and $f_l$ respectively, block residual is $r_l$, then

$$r_l = z_l - b_l \tag{1}$$

$$r_l = z_{l+1} \tag{2}$$

$$\hat{f}_{T+1} = \sum_l f_l \tag{3}$$

The subtraction operation between each baseline block helps mitigate the gradient vanishing problem, allowing the network to become very deep just as the ResNet in [8].

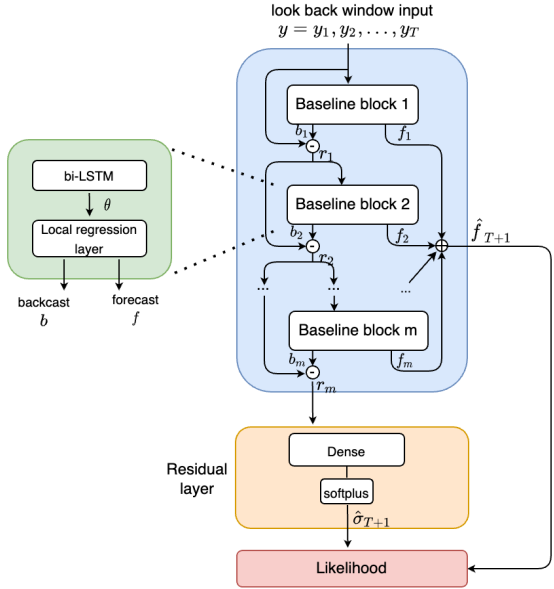

Fig. 1. Network architecture.

Suppose the bi-LSTM network gives forward and reverse outputs $\hat{\theta}_{lf}$ and $\hat{\theta}_{lr}$, the local regression coefficient is then the average of forward and reverse parts:

$$\theta_l = \frac{1}{2}(\hat{\theta}_{lf} + \hat{\theta}_{lr}) \tag{4}$$

The local regression layer will be discussed in section B.

### B. Local regression layer

The classic local regression (aka. LOWESS regression, [13][14]) combines multiple regression models in a k-nearest-neighbor-based meta-model. We integrate the LOWESS idea into neural network. Suppose the degree of the polynomial is $d$, then a standalone polynomial $P_t(x)$ is fitted for the point $t$, $t \in [1, T]$:

$$P_t(x) = \theta_{t,0} + \theta_{t,1}x_t + \theta_{t,2}x_t^2 + \cdots + \theta_{t,d}x_t^d \tag{5}$$

In matrix form:

$$P_t(x) = \theta_t' X_t \tag{6}$$

where $X_t = [1, x_t, x_t^2, \ldots, x_t^d]'$

In contrast to normal least square method, where every sample point plays an equal role in loss, the local regression on $x_t$ weighs more on $x_i, i \in [1, T]$ if $x_i$ is closer to $x_t$. Therefore, a kernel function $K_t(x)$ is chosen to decay the weights. There are two options for $K_t(x)$, one is Gaussian kernel, and the other is tri-cube kernel as described below:

$$K_t(x) = \exp\left(-\frac{(x - x_t)^2}{2H}\right) \tag{7}$$

$$K_{t(x)} = \begin{cases} (1 - |x - x_t|^3)^3, & |x - x_t| \leq H \\ 0, & \text{otherwise} \end{cases} \tag{8}$$

In equation (7) and (8), $H$ is the band width. A smaller $H$ tends to capture the details while a larger H tends to capture the overall trend. $H$ can be kept the same across all baseline blocks. The selection of $H$ is a tradeoff and needs to be cross validated. However, since our network is deep, we set a variable $H$ which means each baseline block has different $H$. In our experiment setup for public datasets, we set larger $H$ for the initial blocks to capture the overall trend, then set smaller $H$ for the latter blocks to capture details. This gives our model powerful capability to model the real baseline. In our empirical studies, the tri-cube kernel achieves even better performance than the Gaussian kernel.

The local regression loss at block $l$, denoated as $\alpha_l$, is the weighted mean squared error described as below:

$$\alpha_l = \frac{1}{T^2} \sum_{i=1}^{T} \sum_{j=1}^{T} K_i(x_j)\left(P_{l,i}(x_j) - z_{l,j}\right)^2 \tag{9}$$

where $z_{l,j}$ $j \in [1, T]$, is the block's input, $P_{l,i}, i \in [1, T]$ is the polynomial of point $x_i$ at block $l$.

The fitted curve of the local regression model between [1, T] becomes the backcast, denoted as $b_l$:

$$b_l = [P_{l,1}(x_1), P_{l,2}(x_2), \ldots, P_{l,T}(x_T)] \tag{10}$$

Meanwhile, we need to keep the backcast smooth, the backcast smooth loss is denoted as $\beta_l$, it's the integral of the curve's second derivatives:

$$\beta_l = \frac{1}{T} \sum_t (b_{l,t+1} - 2b_{l,t} + b_{l,t-1})^2 \tag{11}$$

To forecast the next point after the look-back window for block $l$, denoted as $f_{l,T+1}$, we use the weighted average of the polynomials with each value at $x_{T+1}$:

$$f_{l,T+1} = \frac{\Sigma K_i(x_{T+1})P_{l,i}(x_{T+1})}{\Sigma K_i(x_{T+1})} \quad (12)$$

Ordinary local regression has boundary effect since the kernel weight is not symmetric for points near the boundaries. In the proposed neural network, we can stack many layers of the baseline block to mitigate the boundary effect. It's one of the advantages of our neural network against traditional local regression model.

### C. Residual layer

After baseline extraction blocks, the residual signal is supposed to be white noise. We would like to estimate the variance of the noise and put the mean of the noise close to zero. We utilize a single dense layer, its output passing through a softplus activation is considered as an estimate of standard deviation at time $T+1$, denoted as $\hat{\sigma}_{T+1}$. For noise mean, we compose a mean squared error loss of the residual and minimize it in our network. The residual MSE loss is denoted as:

$$\gamma = \frac{1}{T}\sum_i r_i^2 \quad (13)$$

Where $r_i, i \in [1,T]$ is the network residual vector.

However, the residual does not naturally become a white noise. White noise, by definition, has very small autocorrelation coefficients $\rho_k$, where $k$ is the time lag. Ljung-Box testing[15][16] is the statistical testing for white noise. Then we have the Q statistic in L-B test:

$$Q = T(T+2)\sum_{k=1}^{m}\frac{\rho_k^2}{T-k} \quad (14)$$

Where $m$ is the maximum time lag. In order to turn the residual into white noise, we introduce the Q loss function, which takes the core part from the Q statistic, denoted as:

$$L_Q = \sum_{k=1}^{m}\frac{\rho_k^2}{T-k} \quad (15)$$

To minimize $L_Q$ is to minimize each $\rho_k^2, k \in [1,m]$. We will show the effectiveness of the $L_q$ in the experiments section.

### D. Probabilistic forecasting

So far, the model has provided the point forecast $\hat{f}_{T+1}$, we would like to model the conditional probability distribution of the observed value at $T+1$:

$$P(y_{T+1}|y_1, y_2, \dots, y_T) \quad (16)$$

The point forecast $\hat{f}_{T+1}$ plus the residual noise becomes a Gaussian distribution. Then

$$\hat{f}_{T+1} \sim N(\hat{f}_{T+1}, \hat{\sigma}_{T+1}^2) \quad (17)$$

$$Prob(y_{T+1}) = \frac{1}{\sqrt{2\pi}\hat{\sigma}_{T+1}}\exp\left(-\frac{(\hat{f}_{T+1}-y_{T+1})^2}{2\hat{\sigma}_{T+1}^2}\right) \quad (18)$$

Taking log to the likelihood and then take negative, so to minimize the loss is to maximize the likelihood. The gaussian loss is denoted as $L_g$:

$$L_G = \log\sqrt{2\pi} + \log\hat{\sigma}_{T+1} + \frac{(\hat{f}_{T+1}-y_{T+1})^2}{2\hat{\sigma}_{T+1}^2} \quad (19)$$

At model inference stage, we detect the target value $y_{T+1}$ according to $n$-sigma rule:

$$l_{T+1} = \begin{cases} 1, & |y_{T+1}-\hat{f}_{T+1}| > n\hat{\sigma}_{T+1} \\ 0, & otherwise \end{cases} \quad (20)$$

Where $n$ is the multiplier to the noise standard deviation, it controls the anomaly detection sensitivity.

Finally, the total loss function is the sum of all loss terms, which can be expressed as:

$$L = \sum_l \alpha_l + \sum_l \beta_l + \gamma + L_Q + L_G \quad (21)$$

## III. EXPERIMENTS

In this section, we conduct empirical evaluations for our proposed algorithm on both synthetic and real-world time series. We created two synthetic datasets: one has no anomalies and the other has one anomaly.

### A. Synthetic dataset without anomalies

Figure 2 shows a synthetic time series without anomalies. The time series has a decreasing then increasing trend with white noise added. We use a single baseline block in the model. The baseline block's backcast is shown in yellow, it is smooth and robust to noise. The network residual is supposed to be random noise. We plot the autocorrelation and partial autocorrelation of the residual as Fig 3. The acf and pacf charts are proving the model's hypothesis.

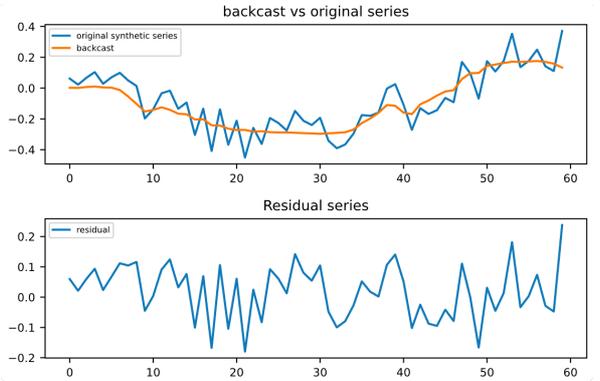

Fig. 2. Baseline block backcast and residual

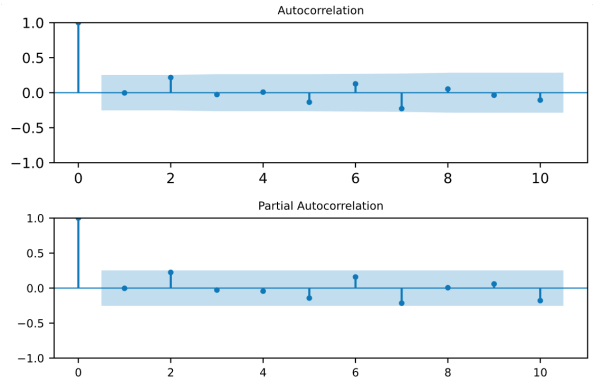

Fig. 3. Network residual analysis

We also conducted the Ljung-Box testing with a null hypothesis that the residual is white noise. The *p*-value is 0.51 which is very high and null hypothesis can be accepted.

### B. Synthetic dataset with anomaly point

Figure 4 shows the synthetic dataset with a high spike anomaly point. The zoomed-in detail between timestep 930 and 1050 shows that the baseline backcast is robust to the anomaly point.

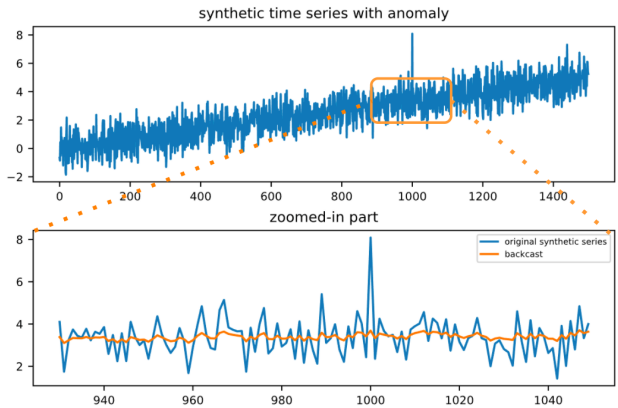

Fig. 4. Synthetic time series with anomaly

### C. Public datasets

To test the effectiveness of our algorithm in anomaly detection, we further conducted experiments on two public datasets:

- Yahoo's Anomaly Detection Dataset [17]. Parts of the dataset are synthetic while others are from real production traffic to some of the Yahoo properties. The time lag between each data point is one hour.
- KPI dataset [18] released by AIOPS data competition. The dataset consists of multiple KPI curves with anomaly labels collected from various Internet Companies, including Sogou, Tencent, eBay, etc. The time lag between each data point is mostly one minute, while some points have 5 minutes interval. The dataset contains two files: one is for training; the other is for testing. The files can be accessed at [20].

The basic facts about the two datasets are listed in Table 1.

Table1: Dataset basic facts

| | Total curves | Min length | Max Length | Total points | Anomaly Points | Anomaly Ratio |
|---|---|---|---|---|---|---|
| Yahoo dataset | 367 | 741 | 1680 | 572966 | 3915 | 0.68% |
| KPI dataset(train) | 29 | 8784 | 146255 | 3004066 | 79554 | 2.65% |
| KPI dataset(test) | 29 | 7578 | 149161 | 2918847 | 54560 | 1.87% |

We run our algorithm in a streaming way, for each point $y_t$, the algorithm gives its forecast $\hat{y}_t$, and its confidence interval $[\hat{y}_{tl}, \hat{y}_{tu}]$, if $y_t$ is outside of the confidence interval, it's judged as anomaly. The confidence interval is based on $n-\sigma$ rule, $n$ is the multiplier of the noise standard deviation. $n$ controls the anomaly detection sensitivity, leading to different levels of recall and precision rates. During training and testing, the algorithm cannot foresee any information beyond $y_t$. Once $y_t$ detection is completed, the algorithm moves on to $y_{t+1}$. Selected results from Yahoo dataset are demonstrated in Fig 5. The gray belt is the confidence interval of $4\sigma$.

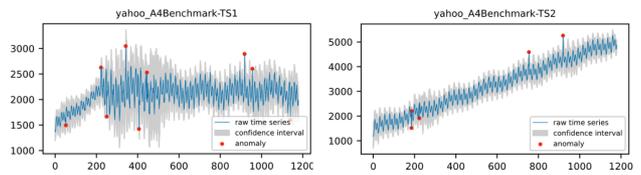

Fig. 5. Anomaly detection examples on Yahoo dataset

Both Yahoo and KPI are labeled datasets. However, in real applications, the point-wise metrics is usually not of interest to the human operators. It is acceptable for an algorithm to trigger an alert for any point in a contiguous anomaly segment if the delay is not too long. Thus, we adopt the same evaluation strategy as in [4] and [11]. The whole segment of continuous anomalies is marked as one positive sample which means no matter how many anomalies have been detected in this segment, only one effective detection will be counted. If any point in an anomaly segment can be detected by the algorithm, and the delay of this point is no more than $k$ from the start point of the anomaly segment, we say this segment is detected correctly. Thus, all points in this segment are treated as correct, and the points outside the anomaly segments are treated as normal. Allowed delay threshold $k$ is set to 3 for hourly data and 7 for minutely data in accordance with [4]. Reference [19] has the evaluation strategy scripts.

| truth | 0 | 0 | **1** | **1** | **1** | 0 | 0 | **1** | **1** | **1** |
|---|---|---|---|---|---|---|---|---|---|---|
| point-wise anomaly | 1 | 0 | 0 | 1 | 1 | 1 | 0 | 0 | 0 | 1 |
| adjusted anomaly | 1 | 0 | 1 | 1 | 1 | 1 | 0 | 0 | 0 | 0 |

Fig. 6. Ilustration of the evaluation strategy, allowed delay $k$=1. The second row is the raw detection result and third row is the adjusted result. In the first continuous anomaly segment, the detection is one step behind, so the whole segment is marked as 1. While for the second continuous anomaly segment, the delay is two time steps, it's not acceptable, so the adjusted label is put to all 0 for this segment.

Table 2: Performance on public datasets

| | Yahoo | | | KPI | | |
|---|---|---|---|---|---|---|
| Model | $F_1$ | Precision | Recall | $F_1$ | Precision | Recall |
| DeepAR | 0.72 | 0.62 | 0.857 | 0.634 | 0.654 | 0.615 |
| SR | 0.563 | 0.647 | 0.598 | 0.622 | 0.647 | 0.598 |
| DBLN | 0.786 | 0.746 | 0.831 | 0.695 | 0.798 | 0.616 |

We compare our results with state-of-the-art unsupervised time series anomaly detection methods. The baseline methods include DeepAR[2] and Microsoft SR[4]. We implement DeepAR and our DBLN by PyTorch. DeepAR has layers of LSTM, connected with a dense layer for mean and a second dense layer for sigma. The two dense layers are in parallel. The size and number of layers of LSTM is subjected to hyper-parameter grid tuning. For Microsoft SR, we directly quote the results reported in [4]. For our DBLN, we implement it with a number of baseline blocks, each block fit a local regression with polynomial degree of 1(linear) or 2(quadratic). For both DeepAR and DBLN, we provide a look-back window of length 120. The test is conducted at a fully streaming way, which means the model cannot foresee any future values. For time series in Yahoo dataset, we take first 500 points in each curve as training set, while the remaining as the test set. For KPI set, there are two data files, one is for training and the other is for testing. The DeepAR and DBLN, we both empirically set $4\sigma$ as threshold for anomaly.

The results comparison is provided in Table 2. All results are collected for the best performance of hyper parameter tuning. It shows our algorithm can achieve superior results to the state-of-art methods. Hyper-parameter tuning is discussed in next section.

## IV. DISCUSSIONS

### A. Ablation study of Q statistic

We analyzed the performance of Q statistic in the loss function as discussed in the residual layer section. We conducted the comparison tests with and without the Q statistic in loss function. The P-R curves on the test sets are shown in Fig 7. For the test with Q statistic in the loss function, the P-R curve is obviously above the one without Q statistic. This proves the effectiveness of our proposed loss function.

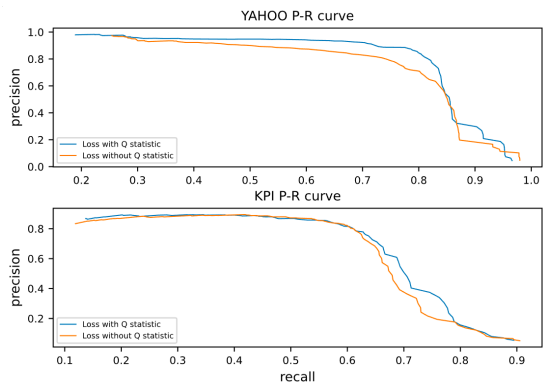

Fig. 7. P-R curve comparision of Q statitics.

### B. Anomaly detection sensivitity

Anomaly is detected by the $n$-sigma rule, $n$ is the multiplier of the noise standard deviation. Different $n$ leads to different results. In real scenarios, human operators can adjust the sensitivity by changing the $n$ value, or set multiple values of $n$ to get results under different sensitivities. In our experiment, we select the $n$ with highest f-score in the validation set. We also plot the P-R curves and f-$n$ curves as in Fig 8.

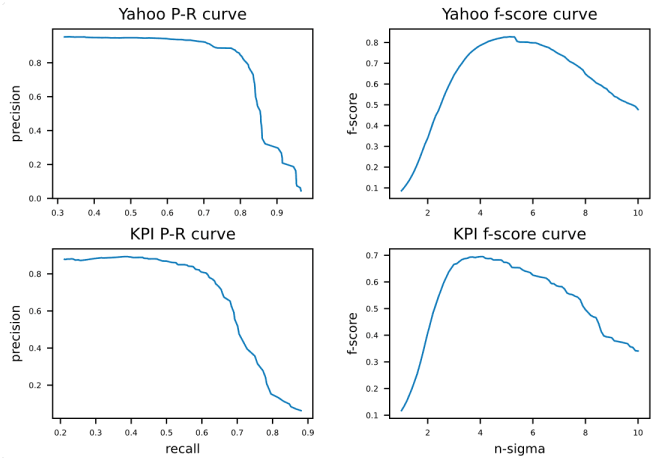

Fig. 8. P-R curves and f-score curve for test sets under different detection sensitivity

### C. Hyper parameter selection

There are several system hyper-parameters in the DBLN network. We conducted a grid-based hyper-parameter search against the validation set. For Yahoo dataset, we take first 0 to 400 points in each curve as training set, 400 to 500 points as the validation set, while the remaining in each curve as the out-of-sample testing set. For KPI dataset, there are two files. We take the last 1000 points in each curve of the training file as the validation dataset. The hyper-parameters include the number of baseline blocks $n$, $H$ in each baseline block, degree of local regression polynomial $d$ and the $l_2$ weight decay penalty. The minimum loss in the validation set is the rule by which we select the best hyper-parameters. The optimal parameters we found for two datasets are listed in Table 3.

Table 3. Hyper-parameter search result

|  | $n$ | $H$ | $l_2$ | d |
|---|---|---|---|---|
| Yahoo | 8 | [8, 8, 8, 8, 6, 6, 6, 6] | 0.001 | 1 |
| KPI | 12 | [10, 10, 10, 10, 8, 8, 8, 8, 6, 6, 6, 6] | 0.001 | 1 |

## V. CONCLUSION

Real-time and interpretability are indispensable in time series anomaly detection scenarios. In this paper, we have integrated the local regression idea into deep neural network. In contrast with N-beats network, where only one low-degree polynomial is fitted, our network models a polynomial regression on each historical point with proper kernel function and bandwidth selected. The proposed Q loss function in the final loss function ensures that useful signal is extracted and residual is white noise. Under this architecture, a robust baseline is extracted and a forecast with probabilistic confidence interval is provided. This leads to clear evidence about anomaly detection. The network performs well in complex trends and noisy scenarios. In the future, we plan to make seasonal pattern modeling more capable in the neural network, so that seasonal factors can also be integrated, and also to apply the model in the time series forecasting scenarios.